\documentclass[letterpaper, 10 pt, conference]{ieeeconf}

\IEEEoverridecommandlockouts                              
\usepackage{lipsum} 

\usepackage{balance}
\usepackage[caption=false]{subfig}
\usepackage[pdftex]{graphicx}

\usepackage{times} 
\usepackage{amsmath} 
\usepackage{amssymb}  
\usepackage{amsfonts}

\usepackage{amsthm}

\theoremstyle{definition}

\newtheorem{remark}{Remark}

\usepackage[vlined,ruled]{algorithm2e}
\SetAlgoSkip{bigskip}

\newcommand{\Kimbla}[0]{\emph{Kimbla}}

\graphicspath{{./pdf/}{./jpeg/}{./figures/}{./svg/}}
\DeclareGraphicsExtensions{.pdf,.jpeg,.png}

\usepackage{cite}

\usepackage{color}              

\title{\LARGE \bf
    Field trial on Ocean Estimation for Multi-Vessel Multi-Float-based Active perception
}

\author{Giovanni D'urso$^1$, James Ju Heon Lee$^1$, Ki Myung Brian Lee$^1$, Jackson Shields$^2$,\\ Brenton Leighton$^2$, Oscar Pizarro$^2$, Chanyeol Yoo$^1$, Robert Fitch$^1$
\thanks{This research is supported by an Australian Government Research Training Program (RTP) Scholarship, the Schmidt Ocean Institute, the University of Technology Sydney and the Australian Centre for Field Robotics.}%
\thanks{$^1$Authors are with the University of Technology Sydney, NSW 2007, Australia {\tt\footnotesize \{Giovanni.Durso, JuHeon.Lee,brian.lee\}@student.uts.edu.au} and {\tt\footnotesize \{Chanyeol.Yoo, Robert.Fitch\}@uts.edu.au}}%
\thanks{$^2$Authors are with the Australian Centre for Field Robotics, University of Sydney, NSW 2006, Australia {\tt\footnotesize \{j.shields, b.leighton, o.pizarro\}@acfr.usyd.edu.au}}%
}

\begin{document}

\maketitle

\begin{abstract}

Marine vehicles have been used for various scientific missions where information over features of interest is collected. In order to maximise efficiency in collecting information over a large search space, we should be able to deploy a large number of autonomous vehicles that make a decision based on the latest understanding of the target feature in the environment. In our previous work, we have presented a hierarchical framework for the \emph{multi-vessel multi-float} (MVMF) problem where surface vessels drop and pick up underactuated floats in a time-minimal way. In this paper, we present the field trial results using the framework with a number of drifters and floats.  We discovered a number of important aspects that need to be considered in the proposed framework, and present the potential approaches to address the challenges.

\end{abstract}

\section{Introduction}
\label{sec:intro}

\begin{figure}[t] 
    \centering
    \includegraphics[width=\columnwidth]{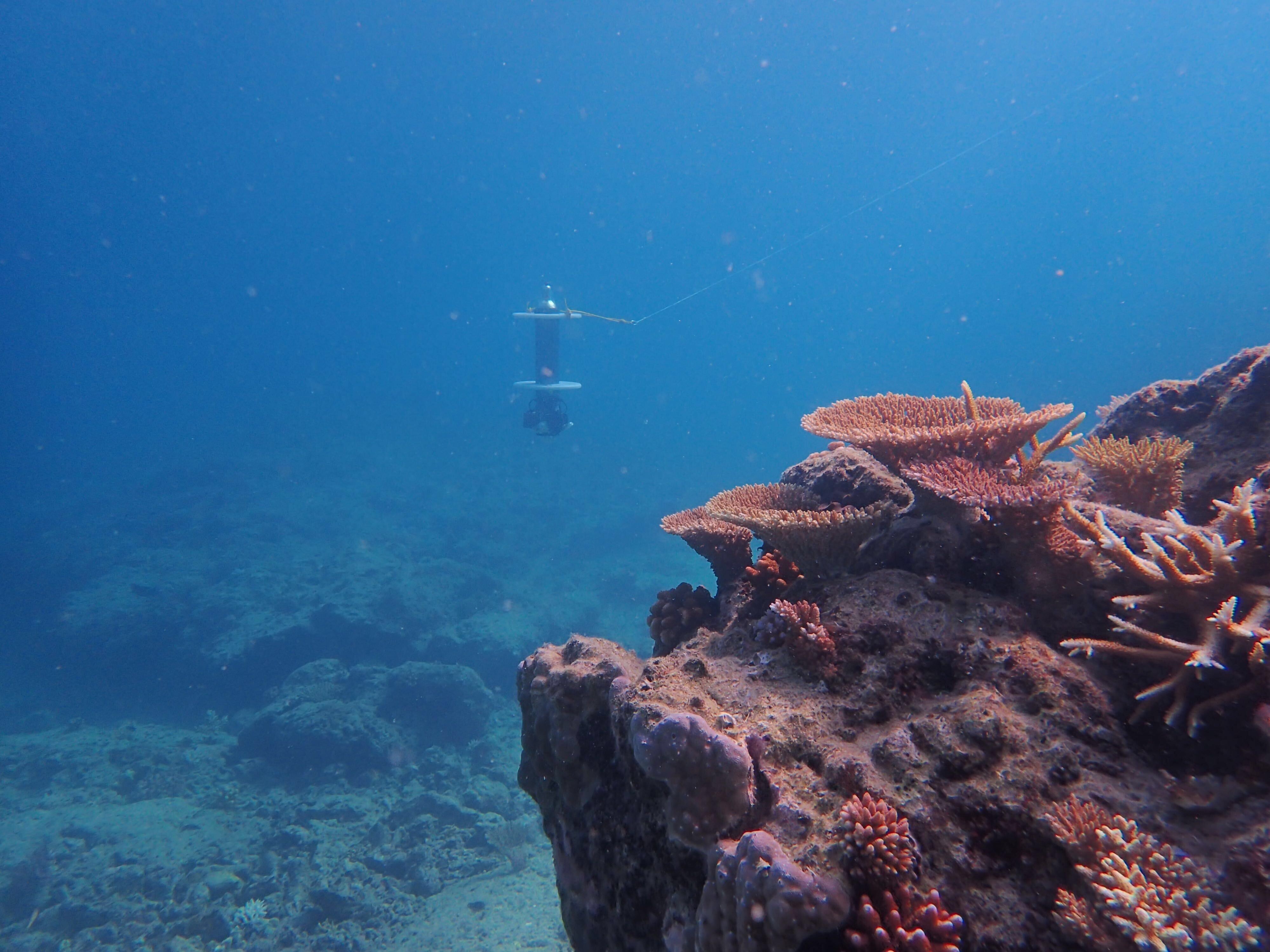}
    \caption{
        Ocean float taking images of underwater features in a field trial for Schmidt Ocean Institute (Lizard Island, Great Barrier Reef, Australia)
    } 
    \label{fig:float_SOI}
\end{figure}

Active perception is a critical component towards achieving persistent autonomy in marine scientific missions such as underwater habitat mapping~\cite{ExplorationHabitat2017,Hyperspectral_mapping2018,JacksonHabitat2020}, environmental monitoring~\cite{Rudnick2004,coralFlorida2006,coralReefReport2011,BenthicRefSite2012}, geological surveying~\cite{Volcano2013,tectonic2014,MidOceanRidges2016}, and plume source detection~\cite{plumeAUV2012,CO2Plume2015,brianPlume2018}.
When the vehicles operate in an oceanic environment, any control or planning done to address active perception must account for time-varying, uncertain oceanic flow field; either by compensating for it~\cite{auv_station_keeping}, or exploiting it~\cite{lee2017energy}.
We are interested in using simple and low-cost marine robots called~\emph{floats} shown in Fig.~\ref{fig:float_SOI} that exploit the ocean current for horizontal movement while the vertical movement is controlled. 

Typical marine missions involve coverage of vast oceanic environment.
In this setting, the under-actuated nature of floats necessitates a \emph{multi-vessel multi-float} (MVMF) system, where autonomous surface vessels (ASVs) judiciously deploy and pick up the floats.  
We have previously studied such a system in the context of visiting points of interest (POIs) while minimising makespan time~\cite{Gio_ICRA2021}. 
In the context of active perception, these POIs can represent scientifically interesting areas for benthic habitat survey. 
To survey a benthic habitat, a typical mission involves a float diving to an imaging altitude and then ``following'' the bottom at a fixed altitude.
Given an oceanic model, the deployment and pick-up of the floats are scheduled such that the horizontal movement induced by the ocean currents maximise coverage of the POIs.

Because the floats are dependent on the surrounding oceanic current, and its sensors have limited field of view, a high-resolution ocean model is necessary to accurately evaluate float trajectories that pass through the POIs.
Such high-resolution ocean forecast data is not readily available in practice.
In addition, solving for MVMF schedule in a time-varying real-world oceanic flow field is difficult, and we have not yet addressed this scope of the problem with our current algorithm.
To address these issues, we proposed in our previous work~\cite{Gio_ICRA2021} that a local flow field estimator can be used to provide ocean data and that the flow field environment was quasi-static; i.e., the flow field can be considered piecewise-constant. 

In this paper, we recount our field trial where we evaluated our assumptions and approaches made in our previous work.
The trial was held along the coast of Jervis Bay, NSW, Australia from the 31st of March to the 1st of April 2021.
During the trial, we demonstrated the applicability of the local flow field estimator alongside our MVMF scheduler, and that the quasi-static assumption does hold for valid planning in a real-world environment.
However, we found that the scope of planning available to us with such assumption and approach was limited to short-term missions. 
We also observed an interesting surface ocean current structure that is beyond the assumptions made for the local flow field estimator.
Based on what we learnt from this trial, we propose new challenges that need to be addressed for a more robust and versatile MVMF scheduler and propose potential approaches to those challenges.




\section{Related work}
\label{sec:relatedWork}

Multi-robot systems have been previously used to improve the efficiency of marine science missions.
The current state-of-the-art systems predominantly use fully actuated AUVs and ASVs, or a combination of both~\cite{science_robotics}.
The sensor data gathered by these vehicles are assimilated using tools such as Gaussian processes~\cite{geoffrey_hollinger,stefanie_kemma,brianPlume2018}, Bayesian neural networks~\cite{JacksonHabitat2020} or reduced-order models~\cite{tahiya_ral}.
The vehicles are coordinated to maximise a performance measure generated from the assimilation result, such as information gain or predictive uncertainty~\cite{geoffrey_hollinger,stefanie_kemma,JacksonHabitat2020}.

We are interested in scaling up such missions to be persistent while ensuring economic viability. 
A promising approach in this direction is to use minimally actuated systems such as underwater gliders~\cite{Rudnick2004} or floats~\cite{argo,ryan_smith}.
In particular, floats are more inexpensive than underwater gliders and have been previously deployed in significant quantities~(approximately 4000 units) for marine science~\cite{argo}.
While these systems enable much longer mission duration (months as opposed to hours~\cite{argo,Rudnick2004,ryan_smith}), they have limited control authority because they are under-actuated.
Our MVMF system provides similar persistence while improving on the control authority through selective pick-up and drop-off with ASVs.

A work closely related to ours is~\cite{ryan_smith}, where a single float is steered deliberately using an accurate, high-resolution model of the ocean currents.
Meanwhile, an open practical challenge is that it is difficult to obtain an ocean current model of sufficient accuracy and resolution (see, e.g.,~\cite{brian2019online,dongsik_chang}).
To this end, we present a practical framework that estimates ocean currents from the execution of plans and generates plans from estimates of ocean currents.

\section{Multi-vessel multi-float problem}
\label{sec:technicalApproach}
We present a brief overview of the hierarchical planner we developed in our previous work~\cite{Gio_ICRA2021}. We then outline the field operations used to test field-applicability of a flow field estimator that will be used to support the operation of the planner.

\subsection{Problem statement}
\label{sec:background}
An intituive description of the MVMF problem, as introduced in our previous work~\cite{Gio_ICRA2021}, is as follows. 
Given a team of surface vessels, floats, the flow field, and the desired locations to visit (points of interests or POIs), we aim to find a sequence of drop-off and pickup actions that allow the underactuated-floats to cover the desired locations.
There are many possible candidate solutions to this problem, and we are interested in finding schedules that visit as many POIs as possible without losing floats by leaving them unattended for long periods after completing their drifts. 

For completeness we provide a short summary of our hierarchical planning method~\cite{Gio_ICRA2021}.
The hierarchical method consists of solving two subproblems:
\begin{enumerate}
    \item Given a time-invariant oceanic flow field~$f_c$, a set of POIs~$\mathbf{Q}$, a set of sampled drop-off actions~$\mathbf{D}$ and its corresponding set of pick-up actions~$\mathbf{P}$, find a set of drop-off actions~$\mathbf{D}_s \subseteq \mathbf{D}$ and its corresponding pick-up actions~$\mathbf{P}_s \subseteq \mathbf{P}$ that observes all unique POIs in~$\mathbf{Q}$.
    \item Given a set of surface vessels~$\mathbf{A}$ and floats~$\mathbf{B}$, and the sets~$\mathbf{D}_s$ and~$\mathbf{P}_s$ found in the previous sub-problem, find the joint schedule~$\pmb{\Phi}$ over all~$\mathbf{D}_s$ and~$\mathbf{P}_s$ that minimises the makespan of the overall mission.
\end{enumerate}
Our method solves the first sub-problem by generating a large number of sampled actions and selecting a subset of pick-up and drop-off actions from these samples using a Monte Carlo Tree Search (MCTS). The second sub-problem is solved by scheduling the selected subset of actions and allocating the actions to the surface vessels using Decentralised Monte Carlo Tree Search (Dec-MCTS). Each surface vessel independently computes its local plan that asymptotically leads to the joint optimal solution.

The largest and untested assumption in our previous work is assuming a fully-known time-invariant flow field. We assumed that this is a reasonable assumption because ocean currents vary slowly and are driven by prevailing meteorological conditions such as tides and wind. Typically, these processes occur on the timescale of hours, while missions can last minutes to few tens of minutes. In other words, it is approximately static.
Another practical limitation of our method is the requirement of a dense flow field model for planning the operations due to the float dynamics and our desired application.
To plan trajectories for the floats that are capable of observing the bathymetric features represented as POIs, and due to the complex bathymetry of reefs and shorelines, we require a spatially dense (vectors at least every meter) over a small-scale (hundreds of meters) model of the flow field. Typical flow field models produced by oceanographic sources are spatially sparse (vectors every hundreds of meters) due to their large scale (hundreds of kilometres). These flow fields are not suitable for our application because the under-actuation of the floats necessitates an accurate knowledge of the flow field for predicting the movement of the system. 
The unavailability of such a model means that we need to produce a local estimate of the flow field rather than rely on existing sources of information in order to apply the planner in practice.

To gain a deeper understanding of the limitations of the fully-known time-invariant flow field assumption, we introduce a process for estimating the flow field locally which is then tested with real-world experiments at Jervis bay. 
To produce the local dense flow field we use the Gaussian process (GP)-based expectation-maximisation (EM) method developed by~\cite{brian2019online} because their approach allows estimation of a dense oceanic flow field from sparse point measurements. The point measurements we use for estimation are the deploy and retrieval locations of the floats or subsampled GPS readings from drifters.
The GP method works by exploiting the incompressibility of ocean currents to deal with the underdetermined nature of the problem. The problem is underdetermined because infinitely many candidate flow fields could have generated the same measurements. By exploiting incompressibility, the search space can be reduced so that a feasible solution can be computed.
There are two open questions that make the applicability of this method to our problem uncertain.
First, the method was tested on sparse GPS point measurements of gliders instead of more dense GPS measurements of drifters. The sparse GP based method might be not applicable for this much data. 
Second, the GP method was shown to work for gliders in deep ocean environments, this field trial takes place in regions of much shallower water (around 10-25 meters depth), near shorelines, and near reefs. It is not yet known if the method will perform as well in this new environment.

\subsection{Sea trial operations}
In this section, we discuss the technical approach that was taken during sea trials to assess the applicability of the flow field GP estimator to real-world oceanic environments similar to our expected operating areas.

To gather this data the drifters or floats need to be spread over the workspace. The size of the workspace and thus needed triangle is limited by both the area that needs to be surveyed by the imaging floats and the communication range of the drifters.
The communication range is important because the drifters communicate their locations blindly without caching. If the vessel is outside the communication range, we partially lose the trajectory information.
Once the drifters are deployed they are left to float along with the current for a fixed period of time then retrieved by the boat. A timely retrieval is important in order to not lose the hardware in the ocean. The selection of the drift length is important because it needs to be long enough to gather sufficient data while not being so long that the data gathered is no longer relevant due to the changing flow field. 
For our tests it was found that around 10-15 minutes of drifting combined with the time taken to deploy and retrieve was sufficient. 

The gathered trajectories from the drifters are then used by the GP based flow-field estimator to build an estimate of what flow field best fits the observed trajectories. 
The GP method is dependant on a set of hyperparameters which can require retuning. The tuning process during sea trials uses a grid search over a range of possible sets of parameters and choosing the best set based on the error between the projected trajectory from the estimated field and the observed field.

\begin{figure}[t] 
    \centering
    \includegraphics[width=1\columnwidth]{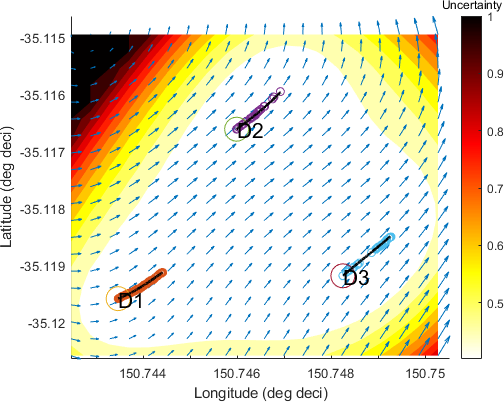}
    \caption{
        Estimated surface ocean current data for a 390m~x~390m oceanic environment. Current estimated from three drifter trajectories deployed for 600 seconds in a triangle formation. The colormap represents the trace covariance. The resulting drift data estimated using GP-based estimator presented in~\cite{brian2019online}. 
    } 
    \vspace{-2ex}
    \label{fig:flowfieldEsti}
\end{figure}
\section{Sea trial result}
\label{sec:results}


Twenty sea trials were conducted over five days at two sites (Murray beach and Nelson beach) inside Jervis Bay, Australia. We present the most representative and interesting data gathered from these trials.

The trials were conducted aboard the MV \emph{Kimbla}, an 18-meter long vessel from which we deployed a heterogeneous team of under-actuated robotic~\emph{floats}~\cite{roman_lagrangian_2011, floats2012,Vision4Float2013} as shown in Fig.~\ref{fig:float_SOI}. These robots were developed by the University of Sydney. The two types of robots we used are a passive \emph{``surface drifter''} type float which has no actuation, and a bottom-following benthic imaging-float which is capable of diving to a controlled depth. For both systems, the motion is mainly driven by the ocean currents which allows interaction with the environment. Both types of robots are equipped with an IMU, GPS, and communicate their state information over LORA. The GPS position information is then filtered by a Kalman filter to reduce noise, 
During these trials, one imaging float and three drifters were utilised to gather the current estimation data.
A visualisation of the drifter deployment for flow field estimation is shown in Fig.~\ref{fig:flowfieldEsti}.


\begin{figure*}[t]
    \centering
    \subfloat[Attempt 1, time: 145 sec] {\includegraphics[width=0.65\columnwidth]{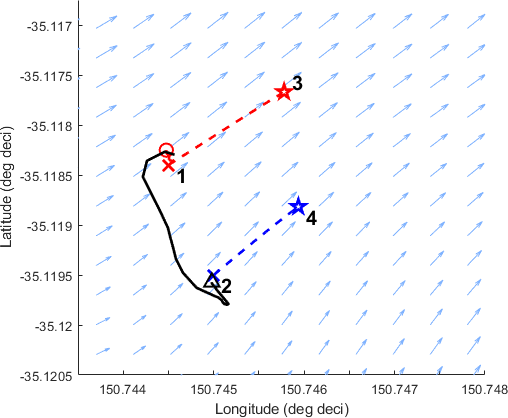} \label{fig:mission1_D2}}
    \subfloat[Attempt 1, time: 660 sec]  {\includegraphics[width=0.65\columnwidth]{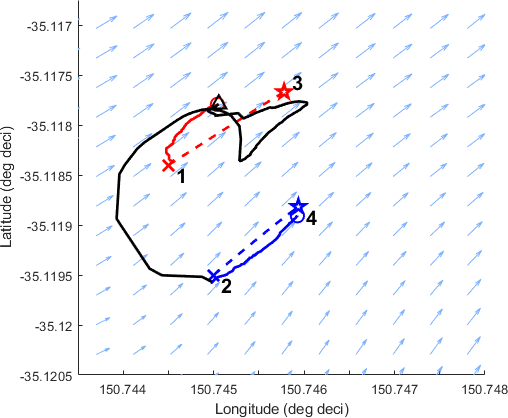} \label{fig:mission1_P1}}
    \subfloat[Attempt 1, time: 790 sec] {\includegraphics[width=0.65\columnwidth]{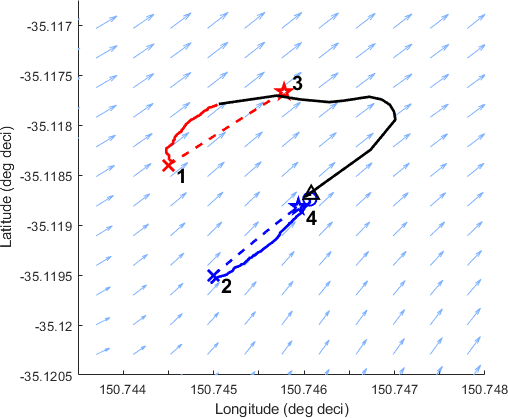} \label{fig:mission1_P2}}
    \caption{
        1 vessel 2 drifter MVMF schedule, planned over the estimated flow field (blue quiver) shown in Fig.~\ref{fig:flowfieldEsti}, executed one hour after the estimation.
        Each subfigure shows a snapshot of the \Kimbla~(triangle) moving from scheduled action to another. For clarity, only the \Kimbla~trajectory (black) associated for that snapshot is shown.
        The sequence of positions the \Kimbla~is to visit are determined \emph{a priori} and are enumerated from 1 to 4; marked with either a drop-off (x) or a pick-up (start) action.
        The actual drifter trajectory (red and blue solid line) is compared against the expected trajectory (red and blue dashed lined). The drifter locations at each snapshot are marked with a circle.
        The \Kimbla~cut through the red drifter path at both (a) and (b) before waiting at the middle of the workspace to ensure communication to all deployed drifters. At time 600~sec, the \Kimbla~moves to position~3 to pickup the drifter before detouring to the actual drifter location then picking up the drifter at position~4. The detour resulted in a tardiness of 60~sec for position~3 and 45~sec for position~4.
    }
    \vspace{-2ex}
    \label{fig:compare1hr}
\end{figure*}

\begin{figure*}[t]
    \centering
    \subfloat[Attempt 2, time: 230 sec] {\includegraphics[width=0.65\columnwidth]{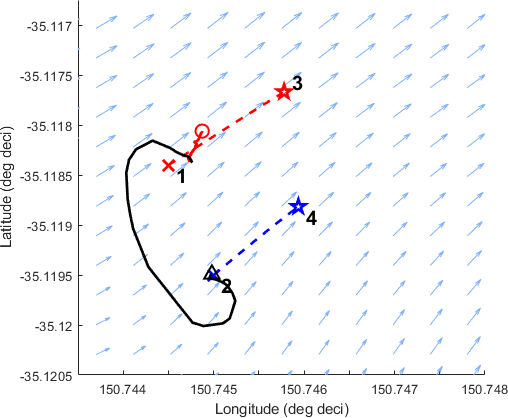} \label{fig:mission2_D2}}
    \subfloat[Attempt 2, time: 680 sec] {\includegraphics[width=0.65\columnwidth]{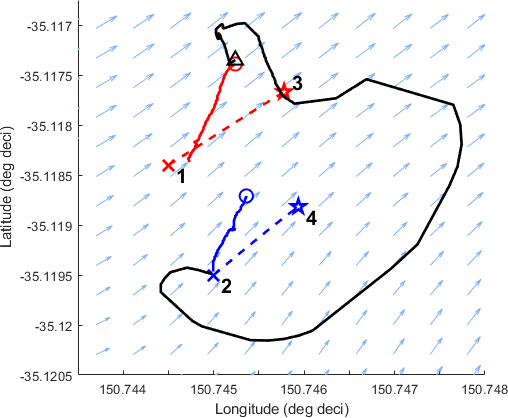} \label{fig:mission2_P1}}
    \subfloat[Attempt 2, time: 855 sec] {\includegraphics[width=0.65\columnwidth]{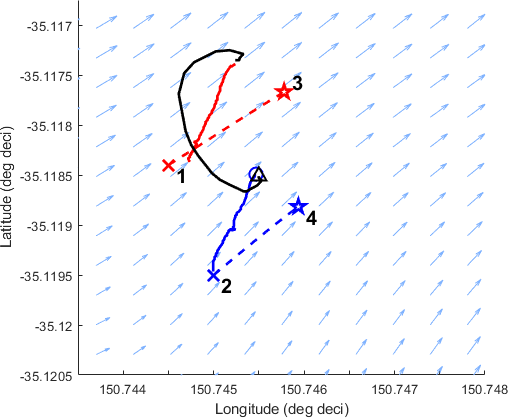} \label{fig:mission2_P2}}
    \caption{
        Same mission as Fig~\ref{fig:compare1hr}, except the schedule was executed two hours after estimation.
        To avoid the effect the \Kimbla~has on the drifters, this mission was executed with slower vessel velocity and having the \Kimbla~travel around the drop-off and pick-up locations. Both drifters however travelled more northerly compared to the expected drifter trajectory, which implies the ocean current have changed. Again, the \Kimbla~visited position~3 at time 600~sec before detouring to the actual drifter location then picking up the drifter at position~4. The detour resulted in a tardiness of 80~sec for position~3 and 25~sec for position~4.
    }
    \vspace{-2ex}
    \label{fig:compare2hr}
\end{figure*}

\subsection{Trial 1 -- Drop-pick evaluation}
We evaluate the consequence of the quasi-static assumption on the MVMF schedule with a simple drop-off/pick-up mission using the \emph{Kimbla} with two drifters.
The \emph{kimbla} operation consists of moving from location to location in the schedule and deploying or retrieving drifters as needed. If a drifter is not at the retrieval location when expected the new retrieval location is set to the current GPS position. The boat then moves to retrieve the drifter at its new location then the mission continues as expected.
We execute the same MVMF schedule twice; one hour after estimation and two hour after estimation.
The evaluation compares the expected drifter trajectory from the schedule with the observed drifter trajectory from both schedule attempts. If the quasi-static assumption holds, then the expected and observed trajectories should be similar.
For the evaluation, we arbitrarily choose two drop-off locations in the workspace, and define the corresponding pick-up locations and expected drifter trajectory by forward integrating across the estimated flow field for 600 seconds.
We also define the action sequence \emph{a priori} as parallel deployment as shown in Fig.~\ref{fig:compare1hr}.


The snapshot of executing the first schedule attempt is shown in Fig.~\ref{fig:compare1hr}.
The \emph{Kimbla} deploys the drifter at point 1 and 2, before waiting at the middle of the work environment to maintain communication between the \emph{Kimbla} and the drifters. After 600 seconds, the \emph{Kimbla} attempts to execute the scheduled pick-up actions while compensating for any discrepancy between the expected and observed result. 
The blue drifter appears to closely follow the expected trajectory.
However, the red drifter trajectory is translated and slower than the expected result.
We suspect that the \emph{Kimbla} may have created wakes near the red drifter that stalled and displaced it. We see that the \emph{Kimbla} cuts across red drifter's expected trajectory in both Fig.~3a and Fig.~3b. 
The delay caused by the wake forces the \emph{Kimbla} to deviate from its schedule to retrieve the drifter, which by then the blue drifter passes over the expected pickup location before it is retrieved.
This result brings up an interesting challenge when planning for MVMF schedule:
\begin{remark} [Disturbance due to surface vessel] \label{remark:wake}
     Recklessly operating the surface vessel near the float robot or its expected path can disrupt the intended float trajectory, which can negatively impact the overall quality of the MVMF schedule.
\end{remark}
This remark applies for both drifter and shallow diving float implementation.
Though the disturbance on the drifters due to the \emph{Kimbla} is more obvious for our small-scale trial, it nonetheless should be avoided for general cases as the ocean current is a chaotic system. In particular, care needs to be taken when deploying drifters for flow field estimation as they deploy around the same time scale as our trial missions.

Snapshot of the second schedule attempt is shown in Fig.~\ref{fig:compare2hr}. 
After learning from our previous attempt, we were cautious to operate the \emph{Kimbla} without creating wake near drifters or its expected path by travelling around the drop and pick location.
While there was no evidence of the \emph{Kimbla} affecting the drifter trajectory, we observed they both drifted more northerly than the expected trajectory.
This implies that the ocean current has changed between the first and second attempt. However, it remained reasonably static between the estimation step and the first attempt as described in:
\begin{remark} [Valid time window for ocean estimation] \label{remark:quasi-static}
    The quasi-static flow field assumption holds true for real-world implementation for about an hour.
\end{remark}

\begin{figure*}[t]
    \centering
    \subfloat[Mission1, time: 1347 sec] {\includegraphics[width=0.65\columnwidth]{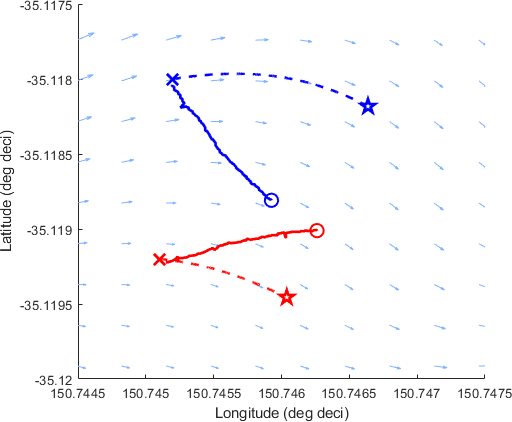} \label{fig:langmuirTraj_1}}
    \subfloat[Mission1, time: 1795 sec] {\includegraphics[width=0.65\columnwidth]{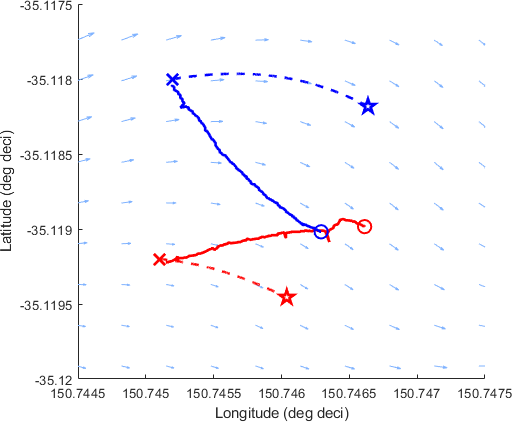} \label{fig:langmuirTraj_2}}
    \subfloat[Mission1, time: 2084 sec] {\includegraphics[width=0.65\columnwidth]{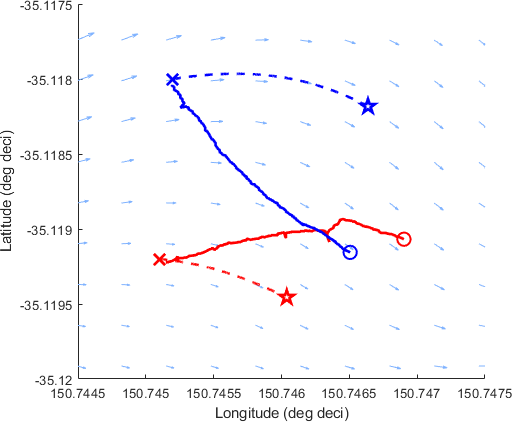} \label{fig:langmuirTraj_end}}
    \caption{
        Snapshot of two drifter trajectory (red and blue) crossing. Expected trajectory and estimated flow field shown for comparison.
    }
    \vspace{-2ex}
    \label{fig:langmuir_circulation_plot}
\end{figure*}
\begin{figure*}[t] 
    \centering
    \includegraphics[width=0.95\textwidth]{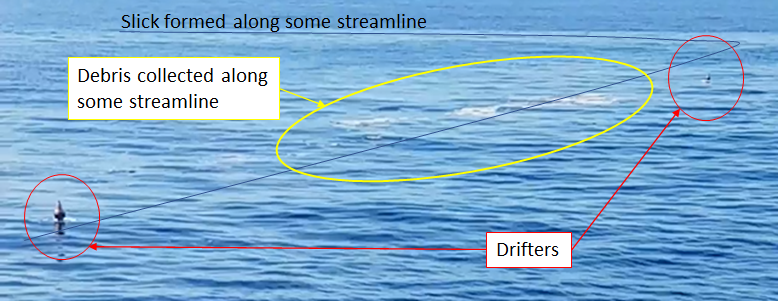}
    \caption{
        Drifters on the ocean surface in an environment consistent with Langmuir Circulations
    } 
    \label{fig:langmuir_circulation}
\end{figure*}
\subsection{Trial 2 -- Langmuir circulation}
During our trial, we have also observed a special flow field structure that contradicts the planar incompressibility assumption made for estimating the flow field.
Two deployed drifters that were expected to move parallel to each other instead moved towards each other until eventually, their trajectories crossed.
The snapshots of this case are shown in Fig~\ref{fig:langmuir_circulation_plot}.

As both drifters were deployed relatively close together mere minutes apart, the trajectories crossing implies that the oceanic streamline are crossed, which is impossible in a planar incompressible flow field.
One hypothesis is that the drifters were sucked into a Langmuir circulation cell.
That is, there is a pair of corkscrew vortex rotating about the direction of the flow that draws nearby water into a stream and pushes it down underwater.
This oceanographic process occurs in the presence of wind shear and when wind and waves move in a similar direction.
We were also able to observe this current phenomenon. Figure~\ref{fig:langmuir_circulation_plot} shows sea debris and slick were collected along a streamline and the floats were observed to be riding these streams.
The existence of such flow structure invalidates the assumption we use for our estimator; that the flow field is planar and incompressible. 
\begin{remark} [Violation to assumption on incompressibility] \label{remark:incompress}
     There exist oceanographic processes that may violate the incompressibility assumption for explicit 2D flow field case near the ocean surface.
\end{remark}


\section{Potential approaches}
\label{sec:discussions}
During our field trial, we have identified various challenges that can be addressed to improve the overall performance of executing a MVMF schedule in a difficult real-world environment.
In this section, we reflect on these challenges and discuss potential approaches to address these challenges.

\subsection{Disturbance due to surface vessel}
From Remark~\ref{remark:wake} and from the second schedule attempt shown in Fig.~3d-f, we need to operate the surface vessels with caution when near a drifter or shallow depth float robot. 
There are several operation-based approaches to mitigate the effect of surface vessels on floats, such as delaying departure to the next position or reducing surface vessel speed when nearby floats. 
One possible algorithmic solution is to plan the surface vessel trajectory such that, topologically, it does not intersect the drifter trajectories nor come near the floats during deployment. Such an approach allows the surface vessel to operate at max capacity while guaranteeing it will not influence the float trajectory.

\subsection{Valid time window for ocean estimation}
As stated in Remark~\ref{remark:quasi-static}, Based on observed data the quasi-static flow field assumption is reasonable for approximately an hour. However, such a window of operation limits the scope and size of the mission due to the slow speed of the floats. In addition, the ocean flow field appears to make sudden changes. That is, we can not use the drift information from the first schedule attempt to predict the flow field for the second schedule attempt.
This challenge motivates a time-varying flow field estimator and planning a MVMF schedule over it, however the time-varying problem is intrinsically difficult. Another approach could be to integrate the estimation step into the MVMF schedule planner and have the system be closed-looped. This approach has the added bonus of being able to autonomously recall floats when the environment becomes too unstable mid-mission.

\subsection{Violation of incompressibility assumption}
From Remark~\ref{remark:incompress}, it is clear that we do not yet fully understand the ocean dynamics enough to make robust flow field estimations. For instance, the GP-based estimator presented in~\cite{brian2019online} does not account for wind, tide, or bathymetry; any of which has an impact on both ocean surface and depth flow field. This challenge motivates a more comprehensive flow field estimator that accounts for other oceanic factors.

The flow field estimator assumes a valid 2D incompressible flow field. External disturbances from wakes or 3D oceanographic phenomena invalidate its assumptions and will lead to incorrect estimations. So care needs to be taken to ensure that the system operates under conditions in which these assumptions hold.
\section{Conclusion}
\label{sec:conclusion}

Our previous work~\cite{Gio_ICRA2021} was validated in a series of field trials with a number of drifters and floats. We have found that there exists several of important extensions to the hierarchical framework in order to improve the overall performance. First, the scheduler should consider the disturbance caused by a surface vessel immediately after deploying drifters and floats. Intuitively, the planner should enforce surface vessel to manoeuvre in a disturbance-minimal manner. Secondly, the oceanic flow changes more rapidly than we anticipated in our previous work. Therefore, there should be a more frequent online update to ocean currents to allow more accurate result. Lastly, our incompressibility assumption on ocean currents may be violated near the surface. Although the assumption still holds true for underwater, such violation should be considered when using drifters and floats near the surface. 

\section*{Acknowledgement}
We would like to thank the crew of the MV \emph{Kimbla} for helping us perform the MVMF field trial. Additional thanks go to the Schmidt Ocean Institute for funding and the University of Technology Sydney and the Australian Centre for Field Robotics for their support in the project.

\balance
\bibliographystyle{IEEEtranS}
\bibliography{SOI_workshop}
\end{document}